# EXTENDING DRT WITH A FOCUSING MECHANISM FOR PRONOMINAL ANAPHORA AND ELLIPSIS RESOLUTION

José Abraços, José Gabriel Lopes - (jea,gpl)@fct.unl.pt

CRIA/UNINOVA, Faculdade de Ciências e Tecnologia, 2825 Monte da Caparica, Portugal

**ABSTRACT**

Cormack (1992) proposed a framework for pronominal anaphora resolution. Her proposal integrates focusing theory (Sidner et al.) and DRT (Kamp and Reyle). We analyzed this methodology and adjusted it to the processing of Portuguese texts. The scope of the framework was widened to cover sentences containing restrictive relative clauses and subject ellipsis. Tests were conceived and applied to probe the adequacy of proposed modifications when dealing with processing of current texts.

## 1. INTRODUCTION

Pronominal anaphora resolution, as part of a more general process of anaphora resolution, is a determinant step in constructing a semantic representation of a text. Although "general cognitive processes DO play a role in establishing anaphoric dependencies (...)" (Kempson, 1990 p.14), inference is, in computational terms, a very expensive process, both for the amount of processing involved and for the extension of the knowledge bases required. Therefore, any system aiming at efficiency in anaphora resolution should minimize the role of inference.

As far as DRT is concerned, the construction rule for pronouns states that the referent introduced by the pronoun should be bound to a *suitable* referent, chosen among those that are accessible (Kamp and Reyle, 1993 p.122). The *accessibility* is based on semantic constraints and is expressed by the structure of DRS representing the text. However the *suitability* of referents is ill-defined.

Another perspective for anaphora resolution is founded on the principle of relevance, i.e. on "the presumption that every utterance is selected to convey the intended interpretation while imposing to the hearer the least amount of processing effort in constructing that interpretation" (Kempson 1990 p.17). Focusing/centering theories (Grosz; Sidner; Brennan, Friedman and Pollard et al.) can be considered as having this perspective. They try to keep track of the focus of attention along the text and bind pronouns preferentially to focused entities. The choice of antecedents is based on pragmatic constraints, which put an ordering on preferences between antecedent candidates.

Cormack proposes the integration of focusing and DRT, "(...) adding semantic constraints to a model of attention in discourse" (Cormack, 1992 p.5). This integration compensates for two shortcomings of DRT: it considers too many possibilities for anaphoric binding and doesn't provide an ordering between antecedent candidates. From the focusing point of view, the addition of semantic constraints, provided by DRT, to the pragmatic ordering further restricts the determination of possible antecedents.

We analyzed Cormack's proposal, and found out that it was lacking some features that we consider more adequate, as it will be shown in the next few sections. Therefore we adapted it, and applied the modified version to the processing of texts written in Portuguese. The scope of those methods was widened to cover sentences containing restrictive relative clauses and subject ellipsis. Tests were conceived and applied to probe the adequacy of proposed modifications when dealing with processing of current texts.

## 2. SIMPLE SENTENCES

### 2.1. Alterations to DRT

Cormack defends that pronouns of the current sentence can only have access to two groups of referents: focused referents and those unfocused ones that were introduced by the preceding sentence. Referents not fitting any of these two groups can be forgotten. Let us look at an example (Cormack, 1992 p.350):

(1a) John took apart the chest of drawers.
(1b) It was full of clothes pegs.

The DRS representing the first sentence will be (focused referents are shown on the left, unfocused ones on the right):

```
<> | < j c >
John(j)
chest_of_drawers(c)
took_apart(j,c)
```

The second sentence introduces another DRS. Anaphors are resolved with referents of previous DRS,

```
<> | < j c >              < c > | < P >
John(j)                   clothes_pegs(P)
chest_of_drawers(c)       full_of(c,P)
took_apart(j,c)
```

and then previous DRS can be "forgotten":

```
< c > | < P >
clothes_pegs(P)
full_of(c,P)
```

Referent *John*, who was introduced by (1a), was only available for anaphor resolution in (1b). Since it was never focused, it is "forgotten". This means that it is no longer included in the referents of the DRS representing the text after processing of the second sentence, becoming unavailable as antecedent candidate for pronouns in following sentences. This claim may seem a little strange if we look at (1c) as an acceptable third sentence:

(1c) *He* didn't like their color.

Two other aspects of Cormack's representation led us to prefer to keep to the original DRT formalism. First, Cormack's representation is too conditioned by pronominal anaphora resolution. Referents that become unavailable for pronominal reference, and are therefore "forgotten", may still be cospecified by definite descriptions. Eliminating them from the representation would be a limit to the possibilities of expanding the system in the future. Second, "forgetting" conditions introduced by previous sentences leads to a situation where the DRS representing the text at a given moment will contain little information about the text, and no information at all about some of the "surviving" referents. For instance, looking at the last DRS presented, we no longer know what entity introduced referent $c$[1].

## 2.2. Focusing algorithms

Most focusing theories keep referents that can be relevant in future anaphora resolution in *focus stores*. Sidner considers two groups of focus stores, which in a very short and simplistic way can be described as:

**those related to agent (AG) role:**
actor focus (AF) - AG of current sentence or previous AF, if current sentence has no AG;
potential actor focus list (PAFL) - other animate referents of current sentence;
actor focus stack (AFS) - previous AFs;

**those related to other thematic roles:**
discourse focus (DF) -
  . DF of previous sentence, if referred with a pronoun in current sentence;
  . referent of the highest ranking pronoun[2] in current sentence;
  . theme, in discourse initial sentences;
potential discourse focus list (PDFL) - referents of current sentence excluding DF;
discourse focus stack (DFS) - previous DFs.

In determining the antecedent of a pronoun, algorithms go through some preliminary considerations (such as *recency rule*) and a basic ordering of focus stores.

### AF - DF distinction

Although taking Sidner's algorithms as a starting point, Cormack renounces the distinction between actor focus and discourse focus, in the final part of her work. The algorithms become more simple but they loose in discriminatory power. This is particularly more significant in a language like Portuguese, where nominals can only be masculine or feminine (not neuter). In a text like

(2a) O João escreveu um livro.
    *John wrote a book. (AF = John, DF = a book)*
(2b) A Maria leu-o.
    *Mary read it.*

eliminating the distinction between AF and DF would lead to *João* (*John*) being proposed as preferred antecedent of the masculine pronoun *o* (*it*). Rejecting this binding would require an appeal to inference, which is something that we want to minimize. Keeping AF - DF distinction will also be significant in dealing with another phenomenon very common in Portuguese: subject (SU) ellipsis.

---

[1] We can, of course, overcome this limitation by creating a text knowledge base where all the restrictions upon referents are present.

[2] see (Sidner, 1979), (Cormack, 1992) for details about this ranking

**Recency rule**

"If the pronoun under consideration occurs in the subject position, and there is an alternate focus list noun phrase which occurs as the last constituent in the previous sentence, test that alternate focus list phrase for co-specification before testing the current focus. (...)" (Sidner, 1979 p.144).

Sidner admits that "the recency rule makes focussing seem somewhat ad hoc" (ibid.), Carter states that "its inclusion in SPAR led to considerable inaccuracy" (Carter 1987 p.114) and Cormack decides to ignore it too (Cormack, 1992 p.54). However, it seems that, in Portuguese, this rule should be considered for pronouns in AG position:

(3a) A Maria$_i$ deu um livro à Ana$_j$.
   *Mary$_i$ gave Ann$_j$ a book.*

If the agent of the next sentence is *Mary* there are two possibilities of pronominalization: the pronoun *ela* (*she*) or the null pronoun ɸ (SU ellipsis). This last option will be preferred:

(3b) ɸ$_i$ comprara-o num leilão.
   *ɸ$_i$ had bought it at an auction.*

But if the agent of the next sentence is *Ann*, the only possibility of pronominalization will be the explicit pronoun *ela* (*she*):

(3b') Ela$_j$ leu-o.
   *She$_j$ read it.*

So the speaker will tend to use a null pronoun in AG position to cospecify the agent of the previous sentence, reserving the explicit pronoun a use that conforms with the recency rule.

**Intrasentential anaphora**

Carter inserts intrasentential candidates (ISC) between current foci and potential foci, in the basic ordering. Cormack distinguishes between *focused ISC* and *remainder of ISC*. In our implementation this distinction seemed unnecessary and we decided to insert ISC after potential foci, in the basic ordering. A special case of ISC is the reflexive pronoun *se* (*himself/herself/itself/ themselves*). We always bind it to the agent of the sentence.

(4) O camelo$_i$ deitou-se$_i$ na areia.
   *The camel$_i$ laid (itself$_i$) down on the sand.*

**Intrasentential cataphora**

In our implementation, syntatic parsing is done according to grammar development formalisms based on barriers, movement and binding (Lopes 1991). It is an extension of the extraposition grammar formalism (Pereira 1981) and allows for movement of constituents of a sentence in a restricted area delimited by barriers. The resulting syntactic tree will always show the internal arguments of the verb on it's right, no matter what positions they had in the original sentence. For instance, the syntatic tree for

(5) Near her, the blond girl saw a man.

will be:

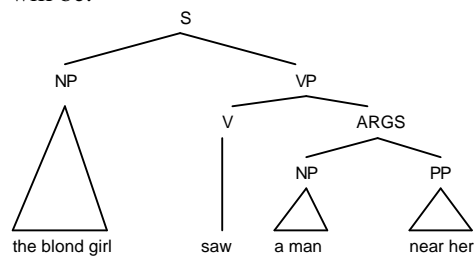

The anaphora resolution process works on the results of the syntatic parser, so this kind of cataphora will be treated as intrasentential anaphora.

**Subject ellipsis**

As mentioned above, this is a very common phenomenon in Portuguese language. Null pronoun in AG position seems to behave differently from one in non-AG position. In the first case it cospecifies AF or a combination of foci including AF:

(6a) A Maria$_i$ decidiu oferecer aquele perfume à Ana.
   *Mary$_i$ decided to offer Ann that perfume.* AF = Mary
(6b) ɸ$_i$ gostava muito dele.
   *ɸ$_i$ liked it very much.*

A null pronoun in non-AG position cospecifies DF or a combination of foci including DF:

(7a) O João poisou o livro$_i$ sobre o piano.
   *John put the book$_i$ on the piano.*   DF = the book
(7b) ɸ$_i$ era grande e pesado.
   *ɸ$_i$ was big and heavy.*

**Ratification procedure**

Both Sidner and Cormack leave all verifications of syntactic agreement and consistency with world knowledge to a ratification procedure, to be applied after completion of focusing process. Efficiency can be improved if inexpensive number and gender agreement and reflexivity verifications are included in the focusing process. Thus, several inadequate candidates can be ruled out without a call to the ratification procedure.

## 3. SENTENCES CONTAINING RESTRICTIVE RELATIVE CLAUSES

Going beyond simple sentences, we widened the scope of the presented methods to include sentences with restrictive relative clauses (for short, we'll just use the form *relative clauses* in the remainder of this paper). Rules for focus movement and referents accessibility were formulated and tests were designed to probe their adequacy. In this section we refer to the results of a questionnaire answered by 40 college students.

**Focus movement**
(8a) O João leu um livro$_i$.
   *John read a book$_i$.*          DF = a book
(8b) O homem$_j$ que o$_i$ escreveu morreu.
   *The man$_j$ who wrote it$_i$ died.*
(8c) Os eruditos enalteceram-no$_{i \vee j}$ ? muito.
   *Erudite people praised him/it$_{i \vee j}$ ? much.*

According to focusing rules, the pronoun in (8c) cospecifies DF of (8b). If pronouns in relative clauses were able to influence focus then (8b) would confirm *a book* as DF and this would be the antecedent of the pronoun in (8c). That doesn't seem to be the case. The intuitively preferred antecedent is *the man*. Examples like this show that pronouns occurring within relative clauses don't seem to influence focus movement. This conclusion was confirmed by 83% of the answers to the above mentioned questionnaire.

**Access of following sentences to relative clause referents**

Referents introduced by the relative clause are accessible but are not preferred to main clause referents. The questionnaire presented the text:

(9) O homem a quem um ladrão roubou o relógio chamou a polícia. Ele ...
   *The man whom a thief stole the watch from called the police. He ...*

58% of the continuations proposed bind the pronoun to the main clause referent *the man* while only 28% indicate binding with the relative clause referent *a thief*.

**Access of the relative clause to main clause referents**
(10) O João deu um livro$_i$ ao aluno que o$_i$ merecia.
   *John gave a book$_i$ to the student who deserved it$_i$.*

Pronouns in the relative clause can cospecify both main clause referents or focus stores. The first situation seems to be preferred except, perhaps, for pronouns in AG position, that show a weak preference (supported by 61% of the answers) for cospecification with AF or a member of PAFL.

**Access of the main clause to relative clause referents**
(11) O homem que escreveu um livro$_i$ deu-o$_i$ à Maria.
   *The man who wrote a book$_i$ gave it$_i$ to Mary.*

Pronouns in the main clause, occurring after the relative clause, can cospecify it's referents. Like Cormack, we consider access to focus stores to be more likely, but this preference was not confirmed by the results of the questionnaire (60% of the answers were against).

**Access of relative clause to relative clause**
(12) O homem que a Maria$_i$ viu escreveu um livro que a$_i$ impressionou.
   *The man who was seen by Mary$_i$ wrote a book that impressed her$_i$.*

Pronouns in the second relative clause can cospecify referents of the first one. However, it seems that main clause referents should be preferred as antecedents. The example used to test this preference was not very clear and so we've got 63% of negative answers.

**Transitive access to a main clause**
(13) A Maria$_i$ casou com o cliente que comprou o livro que ela$_i$ escreveu.
   *Mary$_i$ married the client who bought the book that she$_i$ wrote.*

Pronouns in a nested relative clause can cospecify main clause referents. Preference seems to be given to antecedent candidates of the main clause over those of the nesting relative clause, but this hypothesis was not tested.

**Transitive access to a relative clause**
(14) O cliente que comprou o livro que a empregada$_i$ escreveu casou com ela$_i$.
   *The client who bought the book that was written by the employee$_i$ married her$_i$.*

Pronouns in the main clause can cospecify nested relative clause referents. Candidate antecedents occurring in the nesting relative clause seem to be preferred though. This preference is supported by 75% of the answers.

**Ordering antecedent candidates**

We can summarize this analysis in the following rules for predicting antecedents. These rules were implemented without significant changes to the algorithm established for simple sentences.

<u>Relative clause pronouns:</u>
  AG position:
    φ:  main clause AG
    not null: AF, PAFL, main clause refs., remainder of focus stores
  non-AG position: main clause refs., focus stores
<u>Main clause pronouns:</u>
  Preceding a relative clause: focus stores
  Following a relative clause: idem excluding stacks, relative clause refs., stacks
<u>Following sentence pronouns:</u> main clause refs., relative clause refs., stacks

Nested relative clauses: They have transitive access to main clause refs. Main clause pronouns prefer nesting clause refs. to nested clause ones.

**Relative clauses as conditionals**

Both Kamp (1993 p.81) and Cormack (1992 p.347) propose a "flat" treatment of relative clauses. Both it's referents (with the possible exception of proper names) and conditions are introduced in current DRS.

(15) Jones owns a book which Smith adores.
                                    (Kamp and Reyle, 1993 p.78-83)

| x y z |
|---|
| Jones(x) |
| book(y) |
| Smith(z) |
| z adores y |
| x owns y |

(16) A man who owns a donkey pays.
                                    (Cormack, 1992 p.347)

| <> | <m, d> |
|---|
| man(m) |
| donkey(d) |
| m owns d |
| m pays |

According to Mateus (1979 p.289) the interpretation conveyed by this kind of representation wouldn't be adequate to all kinds of relative clauses in Portuguese, namely those whose verb is in subjunctive mood.

(17) Um agricultor que <u>tenha</u> um burro bate-lhe.
    *A farmer who (subjunctive of <u>to own</u>) a donkey beats it.*

This kind of sentences is associated to non-factual, hypothetical presuppositions and is semantically equivalent to an implication relation between two clauses:

(17') Se um agricultor tem um burro então bate-lhe.
    *If a farmer owns a donkey then he beats it.*

So, our implementation represents this kind of sentences as conditionals:

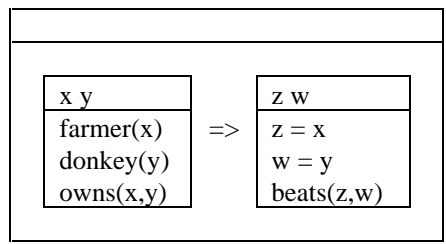

Our rules for anaphora resolution will then be applied as usual, taking in consideration both focusing and semantic (DRT-determined) acessibility constraints.

**4. TESTING**

Tests were conceived with the only purpose of probing the adequacy of proposed modifications. One of the tests, the questionnaire, has already been mentioned. It consisted of two parts. In the first one there were short texts (2-4 sentences) where some referents were introduced. The last sentence was always incomplete and contained a pronoun. The continuation proposed by the student was supposed to show which co-specification he had chosen. Since the evaluation of this part might be influenced by intuition, it was committed to 3 independent evaluators, who were found to agree on 80% of the answers. The second part consisted of texts of the same kind, but where all sentences were complete. The student was asked to identify explicitly the co-specification of a pronoun introduced by the last sentence. The results concerning relative clauses were presented in last section. Recency rule and rules for subject ellipsis were confirmed respectively by 77% and 85% of the answers.

The two other tests consisted of applying the rules for relative clauses to all anaphorae found in current texts, and whose antecedent or anaphor were introduced by a relative clause. The first target text was a novel by a famous Portuguese writer of the end of last century, Eça de Queiroz (1900). The news of a Portuguese news agency (Lusa, 1993) provided 637 kbytes of fresh (June93) raw material for the last test. The rules performed correctly in respectively 96% and 92% of the cases.

## 5. CONCLUSION

We developed and implemented a mechanism for pronominal anaphora resolution, integrating focusing and DRT, and adjusted to Portuguese language processing. Modifications to other authors proposals included recovering AF - DF distinction and recency rule, handling intrasentential anaphora, cataphora, subject ellipsis, restrictive relative clauses and, in particular, those containing subjunctives.

Focusing mechanisms enabled the reduction and ordering of the set of possible antecedents for each anaphor. Final ratification or rejection of each suggested co-specification would require the use of world knowledge and reasoning. That was beyond the aim of this work.

The analysis made for restrictive relative clauses should be extended to other constructions of subordination and coordination, in order to establish more general rules. We believe that many questions raised here might be relevant to processing of other romance languages.